**Global and Local Visual Processing: Influence of Perceptual Field Variables**

Zahra Rezvani[1], Ali Katanforoush[2,3], Richard van Wezel[4], Hamidreza Pouretemad[2*]

[1]School of Computer Science, Institute for Research in Fundamental Sciences (IPM), Tehran, Iran.

[2]Institute for Cognitive and Brain Sciences, Shahid Beheshti University G.C., Tehran, Iran.

[3]Department of Computer Science, Shahid Beheshti University G.C., Tehran, Iran.

[4]Donders Institute for Brain, Cognition and Behavior, Radboud University, Faculty of Science, Nijmegen, the Netherlands

Abstract

The Global Precedence Effect (GPE) suggests that the processing of global properties of a visual stimulus precedes the processing of local properties. The generality of this theory was argued for four decades during different known Perceptual Field Variables. The effect size of various PFVs, regarding the findings during these four decades, were pooled in our recent meta-analysis study. Pursuing the study, in the present paper, we explore the effects of Congruency, Size, and Sparsity and their interaction on global advantage in two different experiments with different task paradigms; Matching judgment and Similarity judgment. Upon results of these experiments, Congruency and Size have significant effects and Sparsity has small effects. Also, the task paradigm and its interaction with other PFVs are shown significant effects in this study, which shows the prominence of the role of task paradigms in evaluating PFVs' effects on GPE. Also, we found that the effects of these parameters were not specific to the special condition that individuals were instructed to retinal stabilize. So, the experiments were more extendible to daily human behavior.

Global and Local Visual Processing: Influence of Perceptual Field Variables on Global Precedence and Global to Local Interference

**Introduction**

The visual world is inherently hierarchical (Navon, 1977; Palmer, 1975). The perception of visual information is initially dominated by the processing of global structure. Global Advantage was first explored by (Navon, 1977) using "hierarchical patterns," in which many small letters (e.g., S or H) were arranged into the shape of a large letter (e.g., an S or H), reflecting distinct local and global levels, respectively. Navon argued that the global shape advantage was reflected by two aspects of performance with compound stimuli: (i) that responses were faster to the global than the local level (global precedence), and (ii) that the global level interfered with responses to the local level (global-to-local interference) ; see (Kimchi, 1992) for a review). Even though both effects reflect the primacy of global shape characteristics in perceptual processing, there is reason to believe that these effects do not reflect the same underlying operation but may actually reflect two distinct mechanisms. In particular, it has been suggested that "sensory mechanisms" (i.e., the magnocellular visual pathway) seems responsible for the Global Precedence Effect, whereas "cognitive mechanisms" (i.e., identification processes) seems responsible for the Interference effect (Poirel, Pineau, & Mellet, 2008).

During four decades, there are many argues about the generality of primacy of whole in different variety of experiments and conditions. This perceptual dynamic is influenced by many factors that can be divided into two major categories: subjective or internal factors, e.g., age, disorder, culture, and the external factors called perceptual field variables (PFVs); e.g., stimulus size, eccentricity, sparsity. Global advantage was examined during different levels of the PFVs across variety of experiments. For more debates about effects of these variables in previous literature, please see the meta-analysis paper. (Rezvani et al., 2020).

In this paper, effects of three important PFVs, namely, Congruency, Size and Sparsity are evaluated on global/local processing, across two different experiments. We explored the effect size of experiment paradigm on global/local processing and show the importance of task design in measuring the effects of PFVs in global/local experiments. On the other hand, interaction effects of all the variables were calculated to show the dependencies between these parameters and their simultaneous effects on global advantage.

In our experiment design subjects were not instructed to fixate on special fixation point and they could have freely eye movements. This feature can bring the experiments closer to the natural conditions of human life.

**Experiment 1**

**Methods**

    **Participants**

Participants were 14 graduated right-handed students at the Radboud university of Nijmegen. All participants were between 20-30 years old. All reported normal or corrected-to-normal vision. Each participant took part in one experimental session of approximately 50- to 60-min duration. Participants attended in the AAQ10 test (Allison, Auyeung, & Baron-Cohen, 2012) without any title to cover the goal of the test and all scored between 0-6 out of 10. This test is a guide for adults with suspected Autism who don't have learning disability. If the individuals score more than 6 out of 10, should consider referring them for a specialist diagnostic assessment. So, this test let us to indicate all are in the typically development group.

**Procedure**

All participants were individually tested in a quiet and darkened room. Participants chins rested on a chin rest located 60 cm from the monitor screen (pixel resolution 1024 by 768, 16-bit color, 100 Hz refresh rate). They were presented with a divided Navon task with one simple and one hierarchical stimuli and were instructed to indicate whether in the predefined target level (global or local) of hierarchical shape is similar to simple stimuli or not, by pressing one of two keyboard buttons. Simple and complex stimuli was randomly situated on the right or left of the screen. Participants were encouraged to respond as fast and accurately as possible. The experiment consisted of two blocks (local and global) of 120 trials with long time distance between them.

In each trial a target stimulus was present after cross fixation. Half of the participants started with the block with target in local level and then completed the block with the global target. The other half of the participants started with the block target in the global level and then completed the block with target in local level. In local block subject instructed to compare between simple shape and local level of complex stimuli and ignore the global level, while in the other block subject instructed to only focus on the global level discarding the other level.

After instruction, all participants completed a practice session with 6 practice trials before commencing with each block of test trials. At the start of each trial, 800 milliseconds fixation cross (0.05') was followed by the appearance of the test stimulus. The stimulus remained on the screen until the participants had responded or three seconds finished. Then a blank page was presented for 200 milliseconds. No feedback was provided. Stimuli were presented and responses were recorded within the Psychtoolbox and Matlab (2017b) environment.

**Stimuli**

Stimulus patterns were automatically generated by manual code written using computer vision system toolbox, Matlab 2017. Every stimulus consists of one simple shape, circle or square in the left or right of screen and one Navon Hierarchical stimulus. The hierarchical stimuli were either large circles made up of small circles or squares or large squares made up of small circles or squares, as shown in Figure1. There were 12 small elements in the parameter of large shapes that situated in the uniform distance from each other.

Size of simple shape were remained fixed in all trials on 3 degree of visual field. While size of hierarchical stimuli was changing between trials randomly from two aspects: global shape size and local shape size. The global shape had one of the five following visual angles: 3', 4.5', 6', 7.5', 9'. The visual angles of local elements of the stimuli are one of 0.075, 0.125 or 0.175 of the global configurations. All stimuli were white on a black background, giving a high contrast.

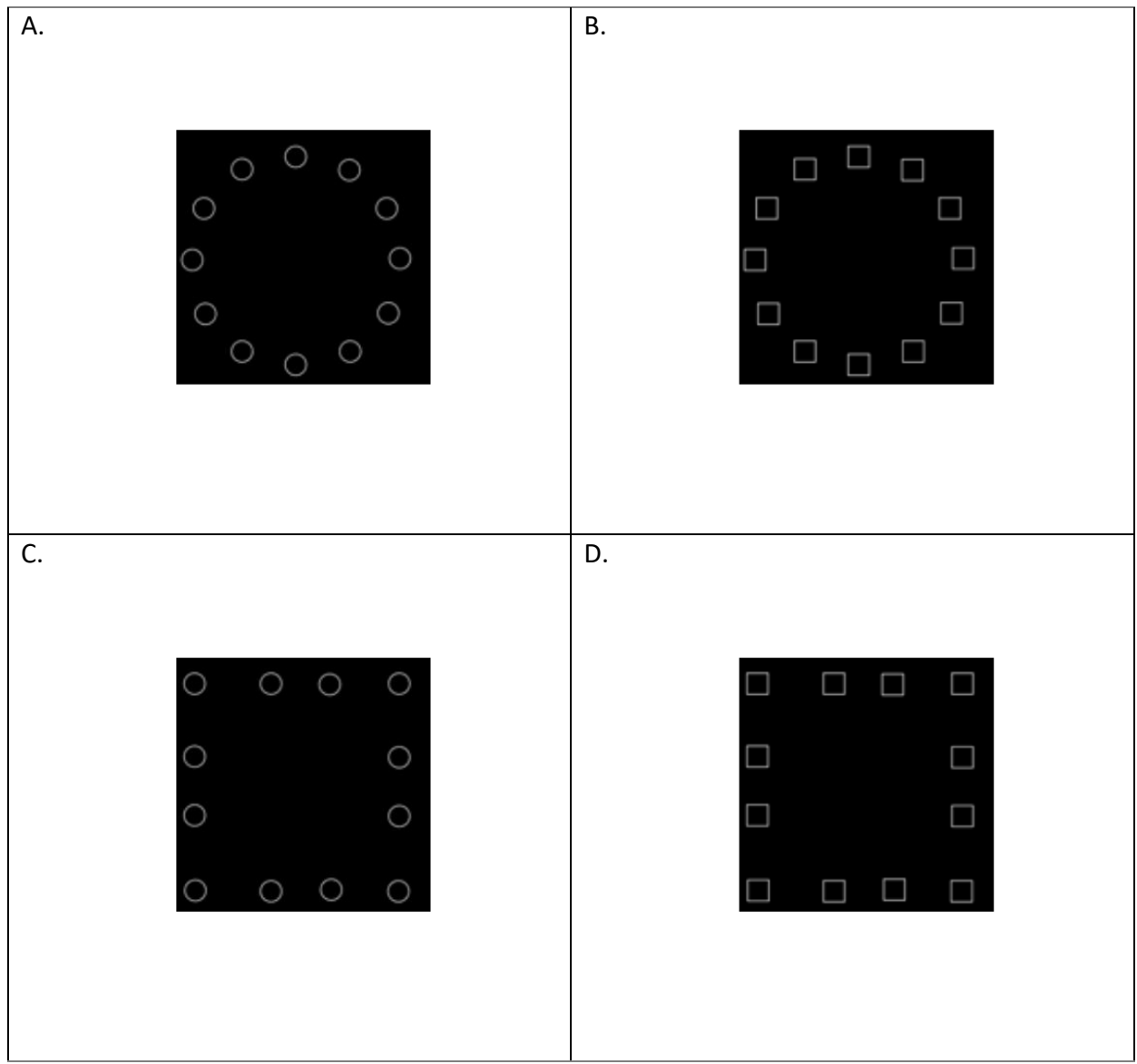

*Figure 1.* The set of stimuli used in the Experiment 1 and 2.

**Results**

Reaction time (RT) of correct response within a range of 200-2000 ms and error rates were analyzed. No outliers needed to be excluded. There was no evidence for speed-accuracy trade off, $R(28)=0.18$, ns. Therefore, in the following section we will report RT analyses only. The RT data were the repeated-measures for Analysis of Variance (ANOVA) with the following factors: Level (global or local target), Congruency (congruent or incongruent), Size (size of hierarchical stimuli global shape in 5 visual angles), Sparsity (relative size of hierarchical stimuli in 3 level). Mean RT for the conditions are displayed in Figure2 (A-C).

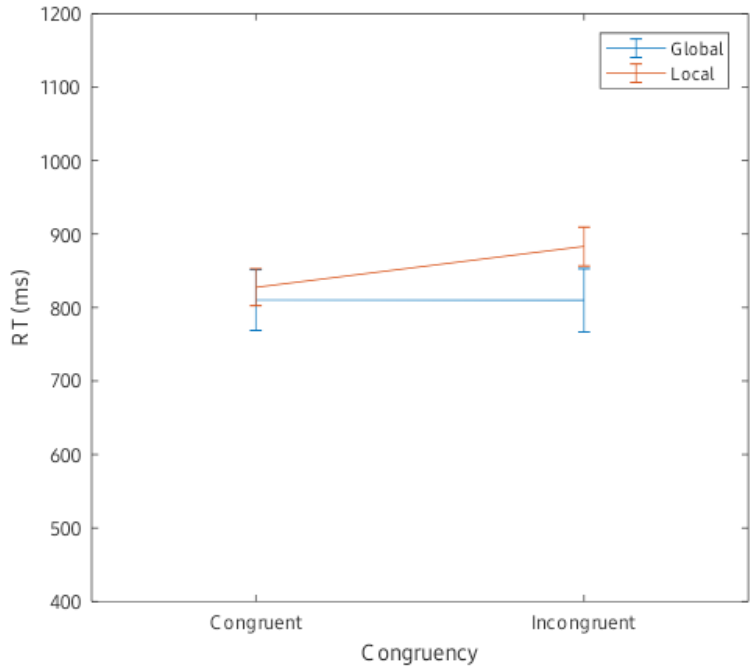

A)

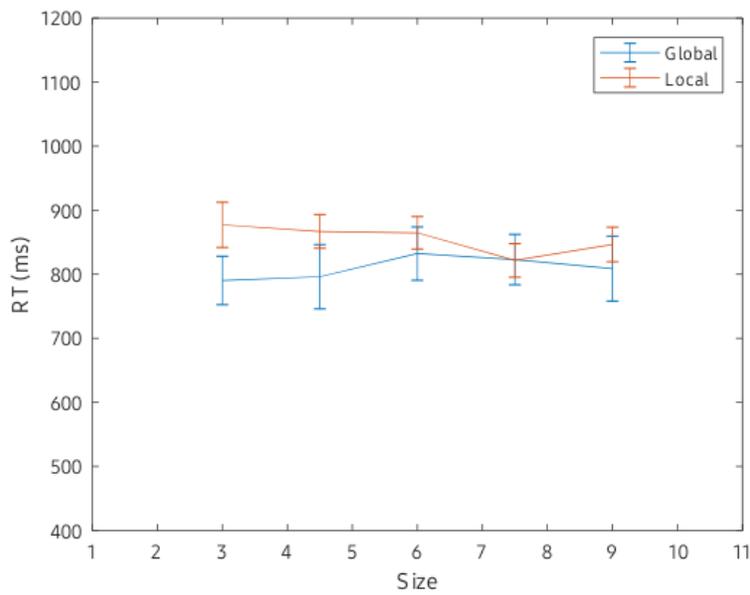

B)

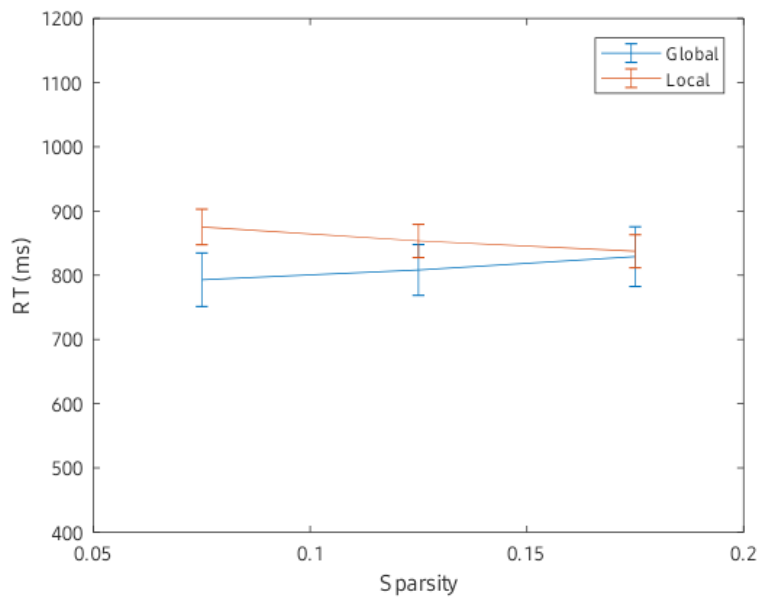

C)

*Figure 2.* Experiment1. The effect of A) Congruency, B) Size and C) Sparsity on global-local processing.

Main effects were observed for first two factors: for Level $F(1,13)= 4.51$, $p<0.05$, with faster responses for global than local level targets and for Congruency $F(1,13)=22.53$, $p<0.001$, with faster responses for congruent stimuli than for incongruent stimuli.

There were two-way interactions between Level and Congruency, $F(1,13)=12.178$, $p<01$, showing that difference in response times between congruency effect was larger for local target than global. It shows the asymmetric interference effect that global information interferes local targets. As well as interaction between Level and Sparsity, $F(2,26)= 6.07$, $p<0.01$, showing that the difference in response time between global and local target in different levels of sparsity. In addition to this, we observed a three-way interaction between Sparsity, Level and Congruency, $F(2,26)=3.33$, $p<0.05$ as well as a three-way interaction between Size, Level and

Congruency, F(4,52)=2.81, p<0.05, showing that the Congruency effect differed between levels of size and sparsity depending on whether participants were asked to respond to the global or local level. Due to the three-way interactions, we then analyzed the data separately for different levels of size and sparsity.

For global part, only effect of Congruency is not significant. But, for local part, effects of Congruency (F(1,13)=27.93, p<0.001), Sparsity (F(2,26)=3.91, p<0.05) are significant.

For incongruent stimulus, effects of Level (F(1,13)=7.48, p<0.01) and its interaction with Size (F(4,52)=2.99, p<0.05) and Sparsity (F(2,26)=8.18, p<0.01) are significant. But, for congruent stimulus, there is no significant effect.

For smallest level of Sparsity, effect of Level (F(1,13)=7.52, p<0.01) and interaction effect of Congruency and Level (F(1,13)=8.52, p<0.01) are significant. For moderate level of sparsity, effects of Level (F(1,13)=7.02, p<0.05) and Congruency (F(1,13)=4.59, p<0.05) and interaction effects of Level and Congruency (F(1,13)=16.02, p<0.01) are significant. And for larger level of Sparsity, there is no significant effect.

For smallest level of Size, effects of Level (F(1,13)=9.18, p<0.01) is significant and Sparsity (F(2,26)=3.91, p<0.05). For second Level of Size effect of Level (F(1,13)=4.91, p<0.05) and interaction effect of Level and Congruency (F(1,13)=24.35, p<0.001) are significant. For third level of Size, effect of Level (F(1,13)=5.49, p<0.05) is significant. For fourth level of size, there is no significant effect. And for last level of Size effects of Congruency (F(1,13)=7.83, p<0.01) and its interaction with Level (F(1,13)=8.52, p<0.01) and Sparsity (F(2,26)=6.40, p<0.01) are significant.

Summarizing Simple effect results shows Global Precedence Effect in except last level of Sparsity and last 2 levels of Size. Also Simple effects of Sparsity show Interference Effects only in moderate level. As well as, simple effects of Size show Interference effect only when the stimulus is on biggest form (9 deg).

**Experiment 2**

**Methods**

    **Participants**

Participants were 14 graduated right-handed students at the Radboud university of Nijmegen. All participants were between 20-30 years old. All reported normal or corrected-to-normal vision. Each participant took part in one experimental session of approximately 50- to 60-min duration. Participants attended in the AAQ10 test without any title to cover the goal of the test and All scored between 0-6 out of 10.

    **Procedure**

As Experiment 1, all participants were individually tested in a quiet and darkened room. Participants chins rested on a chin rest located 60 cm from the monitor screen (pixel resolution 1024 by 768, 16-bit color, 100 Hz refresh rate). They were presented with a divided Navon task with two hierarchical stimuli and were instructed to indicate whether in the predefined target level (global or local) of two hierarchical shapes are similar in that level or not, by pressing one of two keyboard buttons. Participants were encouraged to respond as fast and accurately as possible. The experiment consisted of two blocks (local and global) of 240 trials with long time distance between them.

In each trial a target stimulus was present after cross fixation. Half of the participants started with the block with target in local level and then completed the block with the global target. The other half of the participants started with the block target in the global level and then completed the block with target in local level. In local block subject instructed to compare between simple shape and local level of complex stimuli and ignore the global level, while in the other block subject instructed to only focus on the global level discarding the other level.

After instruction, all participants completed a practice session with 6 practice trials before commencing with each block of test trials. At the start of each trial, 800 milliseconds fixation cross (0.05') was followed by the appearance of the test stimulus. The stimulus remained on the screen until the participants had responded or three seconds finished. Then a blank page was presented for 200 milliseconds. No feedback was provided. Stimuli were presented and responses were recorded within the Psychtoolbox and Matlab (2017b) environment.

**Stimuli**

Stimulus patterns were Automatically generated by manual code written using computer vision system toolbox, Matlab 2017. Every stimulus consists of two Navon Hierarchical stimuli. The hierarchical stimuli were either large circles made up of small circles or squares or large squares made up of small circles or squares, as shown in Figure1. There were 12 small elements in the parameter of large shapes that situated in the uniform distance from each other

Size of hierarchical stimuli was changing between trials randomly from two aspects: global shape size and local shape size. The global shape had one of the five following visual angles: 3', 4.5', 6', 7.5', 9'. The visual angles of local elements of the stimuli are one of 0.075, 0.125 or 0.175 of the global configurations. All stimuli were white on a black background, giving a high contrast.

### Results

Reaction time of correct response within a range of 200-2000 ms and error rates were analyzed. No outliers needed to be excluded. There was no evidence for speed-accuracy trade off, $R(28)=0.02$, ns. Therefore, in the following section we will report RT analyses only as same as experiment1. The RT data were repeated-measures Analysis of Variance (ANOVA) with the following factors: Level (global or local target), Congruency (congruent or incongruent), Size (size of hierarchical stimuli global shape in 5 visual angles), Sparsity (relative size of hierarchical stimuli in 3 level). Mean RT for the conditions are displayed in Figure 3.

Main effects were observed for first three factors: for Level $F(1,13)= 122.22$, $p<0.001$, with high significant faster responses for global than local level targets and for Congruency $F(1,13)=24.24$, $p<0.001$, with faster responses for congruent stimuli than for incongruent stimuli and for Size $F(4, 52)=14.64$, $p<0.001$ with longer reaction times for smallest level of size.

There were two-way interaction between Level and Size, more important than size main effect, $F(4,52)=15.46$, $p<0.001$, showing that the different size effect in response time between global and local level, as well as interaction between Level and Sparsity, $F(2,26)=7.84$, $p<0.01$, showing that the difference in response time between global and local target in different levels of sparsity, and interaction between Congruency and Sparsity, F(2,26)=4.96, p<0.01 demonstrates the difference in response time between global and local targets in different levels of sparsity. However, interactions between Level and Congruency were not significant, showing that difference in response times between congruency effect was not significantly larger for local target than global. It shows the symmetric interference effect that both level information interferes symmetrically. Due to the two-way interactions, we then analyzed the data separately for different levels of size and sparsity.

For global part, only effect of Congruency (F(1,13)=75.50, p<0.001) is significant. But, for local part, effects of Congruency (F(1,13)=5.44, p<0.05), Size (F(4,52)=24.74, p<0.001) and Sparsity (F(2,26)=4.10,p<0.05) all are significant.

For incongruent stimulus, effects of Level (F(1,13)=75.50, p<0.001) and Size (F(4,52)=7.42, p<0.05) and their interaction (F(4,52)=9.11,p<0.001) are significant. But, for congruent stimulus, effects of Level (F(1,13)=100.44, p<0.001), Size(F(4,52)=10.96, p<0.001) and Sparsity(F(2,26)=4.36, p<0.05) and interactions of Level and Size(F(4,52)=5.56, p<0.001) and Level and Sparsity (F(2,26)=3.89, p<0.05), all are significant.

For smallest level of Sparsity, effects of Level (F(1,13)=105.12, p<0.001) and Size (F(4,52)=5.55, p<0.001 ) and interaction effect of Size and Level (F(4,52)=6.50, p<0.001) are significant. For moderate level of sparsity, effects of Level (F(1,13)=128.54, p<0.001) and Congruency (F(1,13)=7.45, p<0.01) and Size (F(4,52)=7.49, p<0.001) and interaction effects of Level and Size (F(4,52)=4.15, p<0.01) are significant. And for larger level of Sparsity, effects of Level (F(1,13)=47.93, p<0.001), Congruency (F(1,13)=15.39, p<0.01) and Size (F(4,52)=4.52, p<0.001) and interaction effect of Level and Size (F(4,52)=5.02, p<0.01) are significant.

For smallest level of Size only effect of Level ($F(1,13)=115.03$, $p<0.001$) is significant. For second Level of Size effect of Level ($F(1,13)=89.81$, $p<0.001$) and interaction effect of Level and Sparsity ($F(2,26)=4.07$, $p<0.05$) are significant. For third level of Size, effects of Level ($F(1,13)=48.25$, $p<0.001$) , Congruency ($F(1,13)=5.37$, $p<0.05$) are significant. For fourth level of Size effect of Level ($F(1,13)=68.26$, $p<0.001$) and interaction effect of Congruency and Sparsity ($F(2,26)=3.51$, $p<0.05$) are significant. And for last level of Size effects of Level ($F(1,13)=38.21$, $p<0.001$) and Congruency ($F(1,13)=12.52$, $p<0.01$) are significant.

Summarizing Simple effect results shows Global Precedence Effect in all levels of Sparsity and Size separately. Also Simple effects of Sparsity show significant effects of Size and interaction of Size with Level in all level of Sparsity. And Interference Effects of when stimulus is not so spars. As well as, simple effect of Size shows Interference effect when the stimulus is not very small (less than 6 deg).

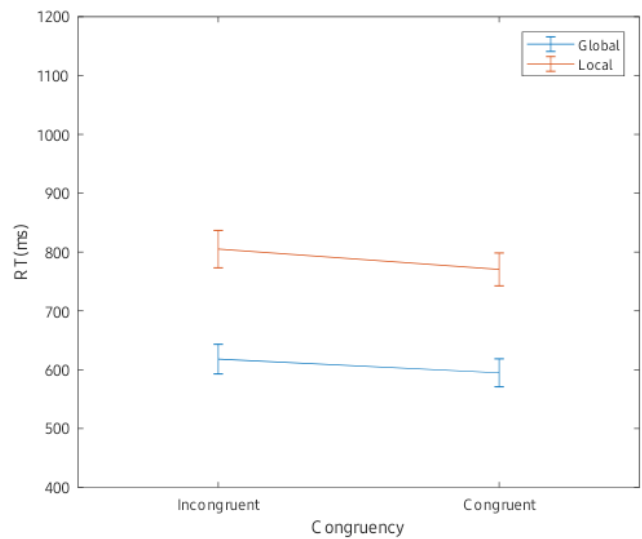

A)

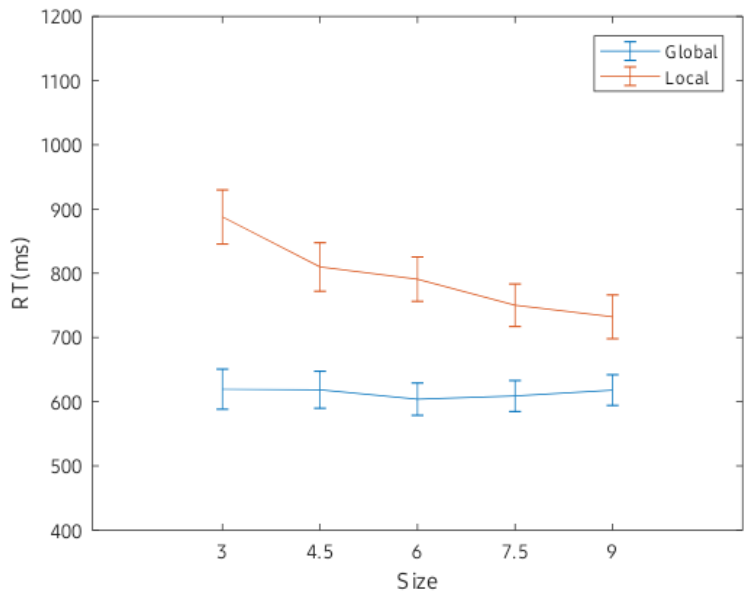

B)

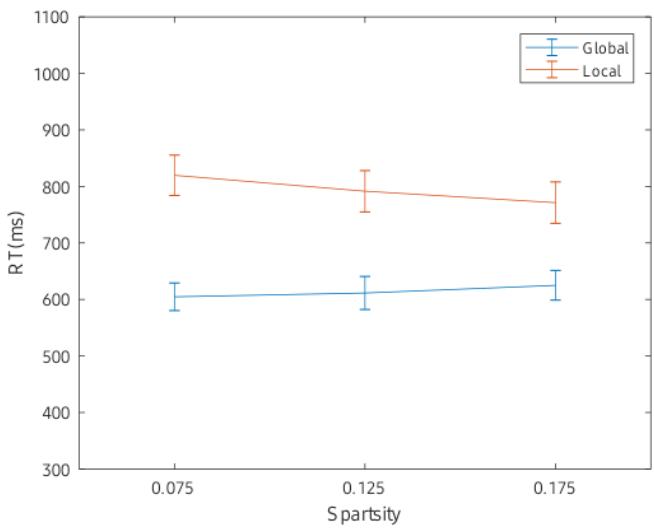

C)

*Figure 3.* Experiment2. The effect of A) Congruency, B) Size and C) Sparsity on global-local processing.

**General Results and Discussion**

Due to comparing this two paradigms and exploring and discussing the effects of every perceptual field variables studied in the above experiments, the RT data were repeated-measures Analysis of Variance (ANOVA)across the two paradigms with the following factors: 1-Paradigm (Matching, Similarity), 2-Level (global or local target), 3-Congruency (congruent or incongruent), 4-Size (size of hierarchical stimuli global shape in 5 visual angles), 5-Sparsity (relative size of hierarchical stimuli in 3 level).

Main effects were observed for first four factors. Statistical results are summarized in Table1 and Table 2. regarding table1, converting effect sizes shows that due to classification and labeling has been done in (Cohen, 1988) Paradigm, Level, Congruency and Size are label as large effect variables and Sparsity is labeled as small effect variable. In our recent study, we have conducted an extensive systematic review and meta-analysis measures pooled effect sizes of most common perceptual field variables (PFVs) in global-local processing (Rezvani et al., 2020). We finally classified the common PFVs in the three levels due to their size of effects. Interestingly our finding in this paper was exactly consistent with the meta-analysis. Congruency and Size were classified as large effect variables and Sparsity was settled in Small effect category as we reported in this paper. For more debates about effects of these variables in previous literature, please see the meta-analysis paper.

More over to these results, we find the paradigm effects as a highly effect variable in global local processing experiments that should be considered in the further experiments about global local processing. Also, regarding to Tables2 and the results of interaction effects of variables reported in the previous sections, we can conclude that interaction effects of variables are very sensitive to task paradigm and design factors.

Another important innovation and new finding of this research was embedded in the instruction of the experiments. In previous experiments subjects was instructed to fixate on special fixation point and prevent them from extra eye movements. But in this study we remove retinally stabilization instruction and found that Global Advantage and the PFV effects on this phenomenon is not restricted to this limitation.

Investigation about Sparsity as a PFV on GPE is arguable. There are many different definitions about this factor in the literature, like number of local shapes, size of them or distance between them or combination of these elements manually and without any standard. In this paper, we use exact and quantifiable measure of local shape size relation with global shape size to measure Sparsity. It has been suggested for further research to explore all the definitions and propose a standard measureable definition based on all the factors.

Table1. The descriptive statistics and results of main effects in both experiments together.

| Factor | Levels | M | SE | df | F | p | $\eta^2$ | Interpretation |
|---|---|---|---|---|---|---|---|---|
| Paradigm | Matching | 817.77 | 29.71 | 1,13 | 32.33 | 0.0001 | 0.71 | **Large effect** |
|  | Similarity | 697.08 | 25.34 |  |  |  |  |  |
| Level | Global | 693.23 | 29.87 | 1,13 | 33.97 | 0.0001 | 0.72 | **Large effect** |
|  | Local | 821.62 | 25.48 |  |  |  |  |  |
| Congruency | Congruent | 743.33 | 24.74 | 1,13 | 70.26 | 0.0001 | 0.84 | **Large effect** |
|  | Incongruent | 771.53 | 26.32 |  |  |  |  |  |
| Size | 3' | 790.10 | 27.68 | 4,52 | 9.45 | 0.0001 | 0.42 | **Large effect** |
| | 4.5' | 760.28 | 28.13 | | | | | |
| | 6' | 762.95 | 25.20 | | | | | |
| | 7.5' | 736.02 | 24.44 | | | | | |
| | 9' | 737.79 | 25.79 | | | | | |
| Sparsity | Sparse | 762.34 | 24.22 | 2,26 | 0.58 | 0.56 | 0.04 | Small effect |
| | Moderate | 755.00 | 27.08 | | | | | |
| | Dense | 754.94 | 26.30 | | | | | |

M: mean; SE: standard error; η2: partial eta squared.

Table2.  Results of interaction effects in both experiments together.

| Factor | df | F | p | η$^2$ | Interpretation |
|---|---|---|---|---|---|
| Paradigm*Level | 1,13 | 10.14 | 0.007 | 0.44 | **Large effect** |
| Paradigm*Congruency | 1,13 | 0.02 | 0.89 | 0.001 | No effect |
| Paradigm*Size | 4,52 | 8.37 | 0.0001 | 0.39 | **Large effect** |
| Paradigm*Sparsity | 2,26 | 0.50 | 0.61 | 0.03 | Small effect |
| Level*Congruency | 1,13 | 9.48 | 0.009 | 0.42 | **Large effect** |
| Level*Size | 4,52 | 10.00 | 0.0001 | 0.43 | **Large effect** |
| Level*Sparsity | 2,26 | 14.88 | 0.0001 | 0.53 | **Large effect** |
| Congruency*Size | 4,52 | 1.54 | 0.20 | 0.11 | Intermediate effect |
| Congruency*Sparsity | 2,26 | 2.48 | 0.10 | 0.16 | **Large effect** |
| Size*Sparsity | 8,104 | 1.81 | 0.83 | 0.12 | Intermediate effect |